\documentclass[review]{elsarticle}

\usepackage{lineno}

\modulolinenumbers[5]

\usepackage{amssymb}
\usepackage{algorithm}
\usepackage{algorithmic}
\usepackage{epsfig}
\usepackage{subfigure}
\usepackage{multirow}
\usepackage{float}
\usepackage{amsmath}
\usepackage{longtable}
\usepackage{booktabs}
\usepackage{subfigure}
\usepackage{fancyhdr}
\usepackage{epstopdf}
\usepackage{ulem}
\usepackage{bm}
\usepackage{amsopn}
\usepackage{amssymb,amsmath,color,times}
\usepackage{amsmath}
\usepackage{amssymb}
\usepackage{amsthm}
\usepackage{bbm}
\usepackage{dsfont}
\usepackage{lineno}
\usepackage{caption}
\usepackage{engord}
\usepackage{lipsum}
\usepackage{float}
\usepackage{graphicx}
\usepackage{color}
\biboptions{numbers,sort&compress}
\journal{Biomedical Signal Processing and Control}
\usepackage[colorlinks,
linkcolor=red,       
anchorcolor=blue,  
citecolor=blue,        
]{hyperref}
\usepackage{cleveref}

\begin{document}
	\begin{frontmatter}
		
		\title{Local-Global Pseudo-label Correction for Source-free Domain Adaptive Medical Image Segmentation}
		
		\author[mymainaddress]{Yanyu Ye}
		\author[secondaddress]{Zhengxi Zhang \footnote{{Yanyu Ye and Zhenxi Zhang contribute equally to this paper and co-share the first authorship of this paper.}}}
		\author[secondaddress]{Chunna Tian\corref{mycorrespondingauthor}}
		\cortext[mycorrespondingauthor]{Corresponding author}
		\ead{chnatian@xidian.edu.cn}
		\author[mymainaddress]{Wei wei}
		\address[mymainaddress]{School of Computer Science, Northwestern Polytechnical University, Xi'an 710129, China}
		\address[secondaddress]{School of Electronic Engineering, Xidian University, Xi'an 710071, China}
		
		\begin{abstract}
			Domain shift is a commonly encountered issue in medical imaging solutions, primarily caused by variations in imaging devices and data sources. To mitigate this problem, unsupervised domain adaptation techniques have been employed. However, concerns regarding patient privacy and potential degradation of image quality have led to an increased focus on source-free domain adaptation. In this study, we address the issue of false labels in self-training based source-free domain adaptive medical image segmentation methods. To correct erroneous pseudo-labels, we propose a novel approach called the local-global pseudo-label correction (LGDA) method for source-free domain adaptive medical image segmentation. Our method consists of two components: An offline local context-based pseudo-label correction method that utilizes local context similarity in image space. And an online global pseudo-label correction method based on class prototypes, which corrects erroneously predicted pseudo-labels by considering the relative distance between pixel-wise feature vectors and prototype vectors. We evaluate the performance of our method on three benchmark fundus image datasets for optic disc and cup segmentation. Our method achieves superior performance compared to the state-of-the-art approaches, even without using of any source data.
		\end{abstract}
		
		
	\end{frontmatter}

	\section{Introduction}
	Medical image segmentation plays a crucial role in image-guided surgery systems, computer-aided diagnosis etc. Recent advancements in deep learning promote the development of automatic medical image segmentation methods that exhibit exceptional performance. Most of these methods rely on abundant labeled data. However, the presence of variations in imaging devices and data sources often leads to domain shifts in real-world medical imaging solutions\cite{vapnik1999overview}. To address this issue, unsupervised domain adaptation (UDA) techniques have been employed, which leverage labeled source domain data and unlabeled target domain data. Nevertheless, in clinical settings, the disclosure of medical imaging data concerns patient privacy. Besides the transmission of data leads to the deterioration of image quality. Source-free domain adaptation is a possible solutions. Research and exploration on source-free domain adaptation (SFDA) methods in medical image segmentation are just started. 
	
	The existing SFDA work can be roughly divided into two categories: source-like domain generation \cite{yang2022source} \cite{ye2021source} and pseudo-label self-training \cite{chen2021source} \cite{xu2022denoising} \cite{vs2022target} \cite{you2021domain}. The former typically involves two steps. Firstly, the distribution information stored in a pre-trained model of the source domain is utilized to generate source-like images. Subsequently, techniques such as Generative Adversarial Networks (GANs) are employed to align the distribution of target domain images with that of the source domain. This alignment process aims to enhance the performance of segmentation in the target domain, particularly when there is a shift in distribution. Pseudo-label self-training methods utilize the source model to generate pseudo-labels for the target domain to train the segmentation models. However, pseudo-labels are often noisy which leads leading to error accumulation problems and negative impacts on target domain segmentation performance.\par
	
	In this paper, we present a novel approach for improving the accuracy and reliability of pseudo-labels in source-free domain adaptation for medical image segmentation. Our method, called local-global pseudo-label correction. To achieve local rectification, we exploit the spatial continuity of images and generate pseudo-labels based on the information ensemble of neighboring pixels, where the label of the current pixel is rectified by the neighbors with high feature similarity. However, relying solely on neighboring pixels within a local region may introduce unreliable pseudo-labels. Therefore, we introduce a global label rectification method based on prototypical feature learning. In our approach, we first divide the target domain dataset into easy and hard samples based on the average entropy at the sample level. We then calculate prototypes for the easy samples and use these prototypes to correct the pseudo-labels by matching pixel-wise features with class prototypes. The prototypes, which aggregate pixel features of the same class, contain common information from different samples. This global rectification complements the local rectification process and ensures that the model learns credible knowledge. By combining local and global rectification techniques, we obtain more reliable and accurate pseudo-labels for training of medical image segmentation in target domian.\par
	
	In summary, our proposed method, called local-global pseudo-label rectification (LGDA), offers a solution for source-free domain adaptation in the context of medical image segmentation. The contributions of our work are as follows:
	\begin{itemize} 		
		\item  We correct pseudo-labels from both local and global perspectives, updating rectified pseudo-labels offline and online during the self-training process, thereby gradually adapting the pre-trained model from the source domain to the target domain data distribution.
		\item We propose an offline local denoising moudle to improve the reliability of pseudo-labels, which employs local context similarity in image space to correct erroneous pseudo-labels pixel by pixel.
		\item We propose an online global denoising moudle to improve the accuracy of pseudo-labels, which split samples into easy ones or hard ones, and corrects erroneously predicted pseudo-labels based on the relative distance between pixel-wise feature vectors and easy samples class prototype vectors.
		\item We evaluate our method on three benchmark fundus image datasets for optic disc and cup segmentation. Without any source data, our method achieves better performance than the state-of-the-art methods.
	\end{itemize}
	
	The rest of this paper is organized as follows. In Section \ref{section2}, we briefly introduce the related work on unsupervised domain adaptation and source-free domain adaptation. In Section \ref{section3}, we introduce the proposed framework in detail. In Section \ref{section4}, we report the experiment settings. In Section \ref{section5}, we analyze the segmentation results. Finally, we summarize this work and give the future work in Section \ref{section7}.

	\label{section:1}
	\section{Related Work}
	We first review related work on unsupervised domain adaptation in subsection \ref{section2_1}. We then describe some related work on unsupervised domain-adaptive image classification in subsection \ref{section2_2}. We present related work on source-free domain adaptation for image segmentation in \ref{section2_3}.
	\label{section2}
	\subsection{Unsupervised domain adaptation}
	\label{section2_1}
	Unsupervised domain adaptation (UDA) refers to the task of adapting a model trained on source domain to target domain without labeled data. Many works have focused on UDA for the classification task, which can be broadly divided into two mianstreams. The first category usually uses Max Mean Distance(MMD), Kullback-Leible Divergence(KL), or others metrics to measure the discrepancy between domains. For example, previous works \cite{long2015learning} \cite{you2021domain} \cite{tzeng2014deep} have used maximum mean difference (MMD) to measure domain similarity. The second category utilizes adversarial training \cite{ganin2016domain} \cite{tzeng2017adversarial}, where a discriminator is employed to distinguish the source and target domains, and a generator is trained to generate domain-invariant features that fool the discriminator, thus achieves domain-adaptive classification.\par
	
	Unsupervised domain adaptive image segmentation is an important application of UDA. Existing UDA methods for image segmentation tasks can be maining categorized into three groups: entropy minimization\cite{vu2019advent}\cite{chen2019domain}, adversarial learning\cite{pan2020unsupervised}\cite{yan2019edge}\cite{zhang2019noise}, and self-training\cite{zou2019confidence}\cite{zou2018unsupervised}. 
	Entropy minimization aims to encourage unambiguous cluster assignments. For instance, Vu et al.\cite{vu2019advent} propose an entropy minimization loss to penalize low-confidence predictions on the target domain and further propose an entropy-based adversarial training method. However, Chen et al.\cite{chen2019domain} have found that entropy minimization suffers from a class imbalance problem and propose a maximum square loss to address this issue. Adversarial learning methods typically employ discriminators to reduce the domain discrepancy between the target and source domains. Pan et al.\cite{pan2020unsupervised} utilize discriminators to minimize inter- and intra-domain gaps, while Yan et al.\cite{yan2019edge} propose an adversarial learning-based approach to learn domain-invariant features in the segmentation task. Additionally, Zhang et al.\cite{zhang2019noise} develop a noise-adaptive generative adversarial network for eye vessel segmentation, where an image-to-image translation generator is designed to map target domain images to the source domain, and a discriminator enforces content similarity between the generated and real images. 
	The self-training method uses the source model to select the pseudo-label with high confidence to supervise the target model training process. Zou et al.\cite{zou2019confidence} proposed a confidence-regularized self-training framework. They treat pseudo-labels as continuous latent variables. In addition, they propose two confidence regularization methods: label regularization and model regularization. Zou et al.\cite{zou2018unsupervised} utilize class-normalized confidence scores to select and generate class-balanced pseudo-labels. However, most of these methods utilize threshold-based filtering of pseudo-labels in the target domain without leveraging associated semantic information to generate more reliable pseudo-labels. In contrast, our method addresses the suppression of noise in target domain pseudo-labels under source-free domain conditions by incorporating local neighboring pixel information and global class prototype knowledge.\par

	\subsection{Source-free domain adaptation for image classication}
	\label{section2_2}
	The previously mentioned UDA methods require the utilization of both source domain data and target domain data simultaneously. However, in practical applications, acquiring the source domain data is challenging due to privacy concerns and data transfer difficulties etc. As a result, the problem of source-free domain adaptation has gained significant attention from medical researchers due to its high clinical value. We review SFDA methods in two applications: classification and segmentation.
	
	Existing source-free domain adaptive image classification works can be broadly divided into two kinds: source-like domain generation and pseudo-label self-training.
	
	Source-like domain generation involves estimating the distribution of source data and generating distribution-similar data of the source domain using target domain data. This allows the SFDA problem to be transformed into a UDA problem. For example, Kurmi et al.\cite{kurmi2021domain} proposed a SDDA model. They first utilized a Generative Adversarial Network (GAN) and a pre-trained classifier to learn the underlying data distribution of the source data. During the training process, random noise is used to generate source-like target images, which are then combined with the target domain data for adaptive training. Li et al.\cite{li2020model} proposed the 3CGAN, which is similar to the method of Kurmi et al.\cite{kurmi2021domain}. However, in the training process, the generation stage and the adaptive stage iterate and collaborate with each other.\par
	
	The pseudo-label self-training methods, employed in the works of Kim et al.\cite{kim2021domain}, Liang et al.\cite{liang2020we}, and Yang et al.\cite{yang2021exploiting}, use a pre-trained model in the source domain to generate pseudo-labels for the target domain. They adopt an iterative training paradigm, where the pseudo-label quality and the segmentation accuracy is gradually improved.\par
	
	\subsection{Source-free domain adaptive for image segmentation}
	\label{section2_3}
	Image segmentation is a pixel-by-pixel classification task. The segmentation task requires the model to have a full understanding of spatial information and global semantic information.The related work on source-free domain adaptive image segmentation can be roughly divided into four categories: source-like domain generation, consistency regularization, prior information guidance introduction and pseudo-label self-training.\par
	
	Specifically,  for source-like domain generation method, Yang et al.\cite{yang2022source}  propose to generate the source-like images by batch normalization constraints through a style loss and a content loss, and further apply mutual Fourier Transform to generate high-quality source-like images. Ye et al.\cite{ye2021source} choose the images with high entropy as the virtual source images and align the distribution by adversarial learning. \par
	
	Secondly, some source-free domain adaptation methods introduce regularization functions. In the literature VS et al.\cite{vs2022target}, Yang et al.\cite{yang2022source}, Ye et al.\cite{ye2021source}, consistency regularization is introduced to the target domain data distribution during the adaptive training of the target domain. In \cite{fleuret2021uncertainty}, dropout is performed on the decoder parameters to obtain different inputs, and then consistent regularization is applied to multiple predictions to train the network.\par
	
	Thirdly, other scholars have used the anatomical prior information of the tissue to be segmented to regulate the source-free domain adaptation process. Bateson et al.\cite{bateson2020source} introduced an auxiliary network to predict the class ratio of objects inspired by anatomical knowledge. In the domain adaptation stage, KL divergence is used to measure the difference between the target class ratio and prior knowledge in the target domain segmentation result. This difference is used to train the network to adapt the source domain pre-trained model to the target domain data distribution. \par
	
	Last, some methods utilize the source model to generate pseudo labels and finetune the source model. To filter out some noisy components in pseudo-labels, Chen et al.\cite{chen2021source} enhance the quality of pseudo-labels by introducing complementary pixel-level and class-level pseudo-label denoising methods, which incorporates uncertainty estimation and prototypical feature learning to reduce noisy components in pseudo-labels. VS et al.\cite{vs2022target} propose a two-stage method including a target-specific adaptation stage and a task-specific adaptation stage. The target-specific adaptation employs an ensemble entropy minimization of multiple pseudo-labels. The false-negative predictions in pseudo-labels are filtered out by selective voting. In the task-specific adaptation stage, strong-to-weak consistency learning is introduced by the teacher-student network. \par 
	
	Our method belongs to pseudo-label guided self training. Notablly,  we propose an offline local label correction method and an online global pseudo-label correction method in LGDA. Offline local pseudo-label correction can provide reliable initial pseudo-labels for target domain retraining. Online dynamic global pseudo-label correction can further improve the accuracy of pseudo-label supervision information with the improvement of model performance. \par
	
	\section{Methodology}
	\label{section3}
	\begin{figure}[h]
		\centering\includegraphics[scale=0.3]{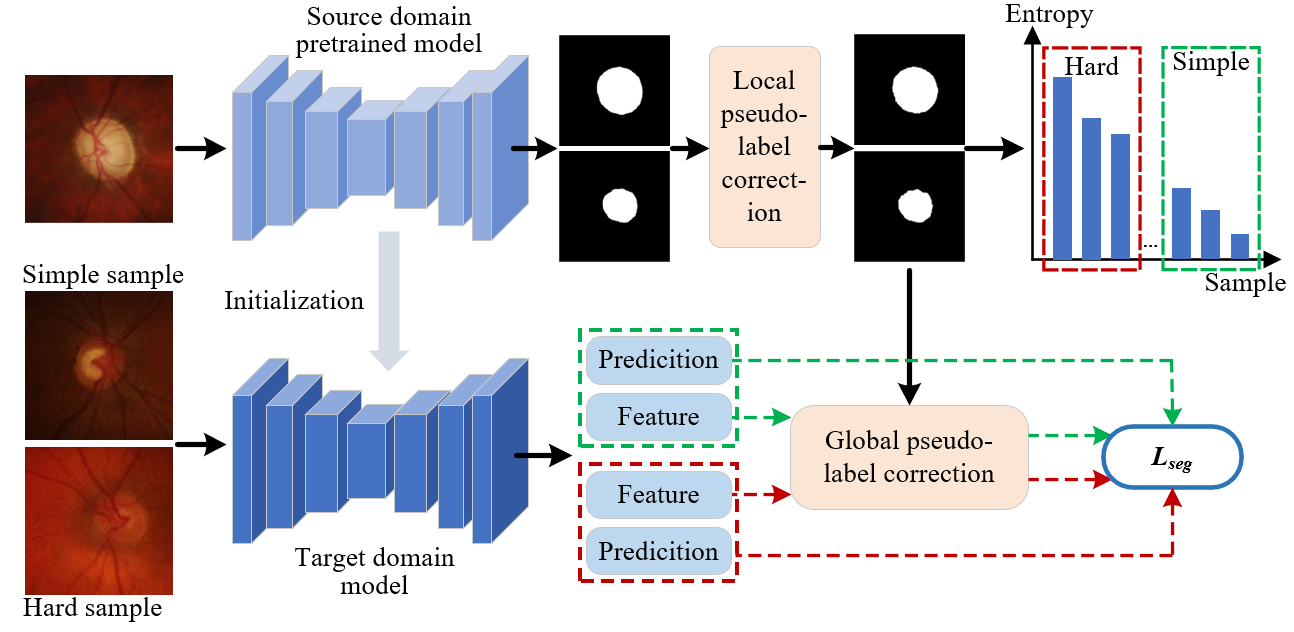}\caption{Source-free domain adaptation medical image segmentation method based on the local-global pseudo-label correction.}\label{method}
	\end{figure}
	\subsection{Overview}
	We denote the source domain training set as ${D_S} = \{ (x_i^S,y_i^S) \in ({X_S},{Y_S})\} _{i = 1}^{{N_1}}$, where $x_i^S$ is the $i$-th original image in the source domain, $y_i^S \in {[0,1]^{H \times W \times C}}$ is the ground truth corresponding to the $i$-th image. $H$ and $W$ are the height and width of the image and ground truth, and $C$ represents the number of categories. Likewise, the unlabeled target domain image is defined as ${D_T} = \{ x_i^T \in ({X_T})\} _{i = 1}^{{N_2}}$. The source domain model pre-trained on the source domain dataset is written as ${\phi _S}(\cdot)$, and the adaptation target domain model is defined as ${\phi _T}(\cdot)$. Figure \ref{method} illustrates our SFDA method with local-global pseudo-label correction.
	
	\subsection{Segmentation Network}
\begin{figure}[h]
	\centering\includegraphics[scale=0.65]{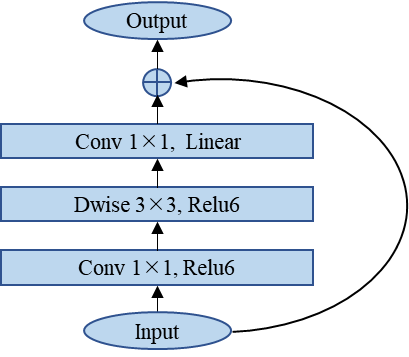}\caption{Bottleneck Inverted Residual block}\label{bottle}
\end{figure}

	We employed the DeepLabv3+\cite{chen2018encoder} network based on MobileNet v2 \cite{chollet2017xception} as our backbone network. MobileNet v2 incorporates an innovative Inverted Residual structure that effectively transforms the input into the desired output by expanding and subsequently reducing the dimensions. The schematic diagram of the bottleneck inverted residual structure is shown in Figure \ref{bottle}. Table \ref{MobileNetlayer} illustrates the network architecture of MobileNet v2, where t represents the channel expansion ratio, i.e., the expansion factor; $c$ denotes the number of channels; $n$ indicates the repetition count of the current bottleneck inverted residual structure, and $s$ represents the stride. 
	
	\begin{table}[H]
		\renewcommand{\arraystretch}{1}
		\centering
		\caption{MobileNet v2 network}
		\label{MobileNetlayer}
		\begin{tabular}{cccccc}
			\hline
			input       & Operation          & $t$ & $c$    & $n$ & $s$ \\
			\hline
			$224^2\times3$   & conv2d      & - & 32   & 1 & 2 \\
			$112^2\times32$  & bottleneck  & 1 & 16   & 1 & 1 \\
			$112^2\times16$  & bottleneck  & 6 & 24   & 2 & 2 \\
			$56^2\times24$   & bottleneck  & 6 & 32   & 3 & 2 \\
			$28^2\times32$   & bottleneck  & 6 & 64   & 4 & 2 \\
			$14^2\times64$   & bottleneck  & 6 & 96   & 3 & 1 \\
			$14^2\times96$   & bottleneck  & 6 & 162  & 3 & 2 \\
			$7^2\times160$   & bottleneck  & 6 & 320  & 1 & 1 \\
			$7^2\times320$   & conv2d $1\times1$  & - & 1280 & 1 & 1 \\
			$7^2\times1280$  & avgpool $7\times7$ & - & -    & 1 & - \\
			$1\times1\times1280$ & conv2d $1\times1$  & - & $k$    & - &  \\ \hline
		\end{tabular}
	\end{table}

	\subsection{Local Denoising Module}
	\label{modelstatus}
	Due to the domain shift between different datasets, the pseudo-labels generated by the source model are bound to contain erroneous predictions. Incorporating these incorrect pseudo-labels into the model adaptation process can lead to the accumulation of errors, thereby hindering the enhancement of the model's segmentation performance in the target domain. To address this issue, we propose leveraging the local spatial continuity of images to rectify the pseudo-labels assigned to individual pixels. It is important to highlight that the module utilizes an offline mode to rectify pseudo-labels, thereby providing reliable initial pseudo-labels for target domain retraining and effectively reducing error accumulation. We elaborate on the Local denoising module in detail in the following. We elaborate on the local denoising module in detail in the following.\par
	
	\begin{figure}[h]
		\centering\includegraphics[scale=0.32]{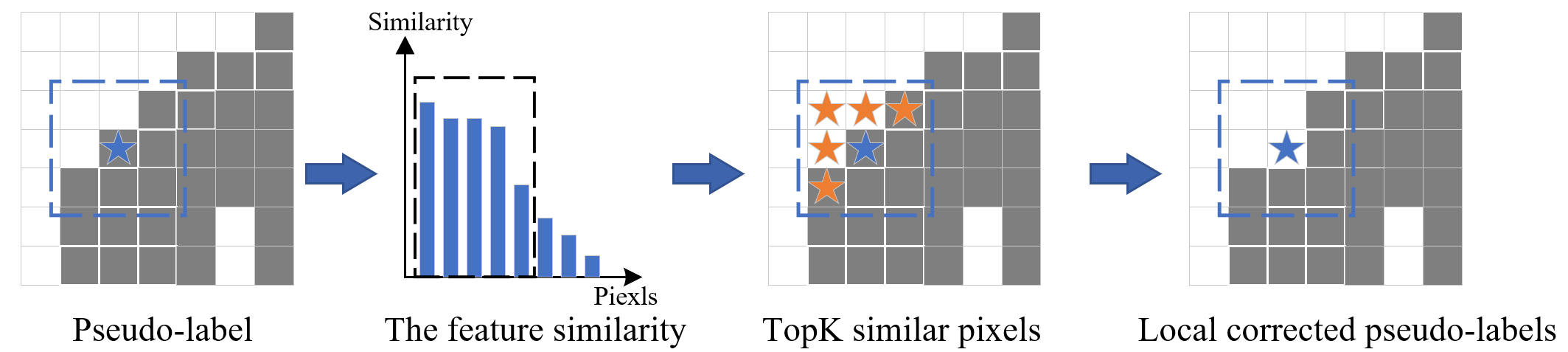}\caption{Local pseudo-label correction}\label{local}
	\end{figure}
	
	First, given an input image from the target domain, we perform dropout sampling with $Num$ forward passes. Then compute the mean of $Num$ predictions as features and pseudo-labels for the target domain images, which is formulated as:\par
	
	\begin{equation}
		f_i^{S \rightarrow T},p_i^{S \rightarrow T} = mean({\phi _S}{({x_i}^T)_{Num = 1,2,...,10}})\label{eq1}
	\end{equation}
	
	\noindent where $f_i^{S \rightarrow T} \in { \mathbb{R}^{{H_1} \times {W_1} \times {C_1}}}$ is the feature vector of the $i$-th image in the target domain. $p_i^{S \rightarrow T} \in {{\mathbb{R}}^{{H_2} \times {W_2} \times {C_2}}}$ is the pseudo-label of the $i$-th image in the target domain. ${H_1}=128$, ${W_1}=128$, ${H_2}=512$ and ${W_2}=512$ are the height and width of the feature matrix and pseudo-label. ${C_1} = 512$ is the number of channels of the feature. ${C_2}$ is the number of segmentation categories, the size of ${C_2}$ is 2 in the optic cup and optic disc segmentation task. The first channel is the predicted probability for the cup class, The second channel is the predicted probability for the disc class. To generate the feature matrix whose size is the same as the pseudo-labels, we recover the feature dimension using the bilinear interpolation method.\par
	
	We use the feature vector of the pixel to calculate the similarity between pixel $j$ and its surrounding pixels, which is formulated as follows:\par
	\begin{equation}
		s_i^{S \rightarrow T}(j,j + o) =  < f_i^{S \rightarrow T}(j),f_i^{S \rightarrow T}(j + o) >\label{eq2}
	\end{equation}
	where $f_i^{S \rightarrow T}(j)$ is the feature vector of pixel $j$. $f_i^{S \rightarrow T}(j + o)$ is the feature vector of pixel $j+o$, and $o \in N \times N$ is the neighborhood of pixel $j$. $s_i^{S \rightarrow T}(j,j + o)$ is the similarity between $j$ and $j+o$. $<\cdot>$is the similarity measure function, and we choose the cosine function as the similarity measure function.\par

	\begin{equation}
		s_i^{S -  > T}(j,j + o) = \frac{{f_i  ^{S -  > T}(j) \cdot f_i^{S -  > T}(j+o)}}{{|f_i^{S -  > T}(j)| \times |f_i^{S -  > T}(j + o)|}}
	\end{equation}
	
	After obtaining the similarity, we use the TopK algorithm to select the pseudo-labels of the first $K$ pixels to correct the label value of pixel $i$. The rectified pseudo-label of image $i$ is calculated as,
	\begin{equation}
		p_i^{S \rightarrow T}(j) = \sum\limits_{o \in A} {p_i^{S \rightarrow T}(o)} /K,A = topK(s_i^{S \rightarrow T})
	\end{equation}

	\subsection{Global Denoising Module}
	The local denoising module is designed to utilize the local context information among neighboring pixels within an image to correct the pseudo-labels.  However, it is still possible for some pseudo-labels of neighboring pixels to be inaccurately predicted.  Consequently, the local denoising module alone is insufficient for effectively filtering out the noisy components in the original pseudo-labels.  To overcome this limitation, we propose an online pseudo-label rectification method that incorporates class prototypes containing global contextual information to correct the pseudo-labels during the training process.  To obtain accurate class prototype vectors, we first divide the samples in the target domain training set into easy and hard samples.  Subsequently, pseudo-label correction is then performed by comparing the relative distances of pixel-wise features to foreground prototypes and background prototypes.  Furthermore, to prevent the model from overfitting to the initial pseudo-labels and producing sub-optimal performance, we update the pseudo-labels in an online manner.  The proposed global denoising module is described in detail below.
	
	\noindent{\textbf{Easy and Hard Sample Division:}}
	Firstly,  the pixel-level predicted probability $p(x_i^T(j))$ and the feature vector $f_i^T(j)$ of pixel $j$ from the image $i$ in the target domain are generated by the pre-trained source domain model. Then, we use Eq.(\ref{entropy}) to calculate the overall entropy of each image.\par
	\begin{equation}
		\label{entropy}
		H(x_i^T) = \sum\limits_i {P(x_i^T(j))I(x_i^T(j))}  = \sum\limits_i {p_i^{S -  > T}(j){{\log }_b}} p_i^{S -  > T}(j)
	\end{equation}
	
	Then, we sort the all images in the target domain based on $H(x_i^T)$, which is formulated as:
	\begin{equation}
		img\_list = sort(H(x_i^T)),x_i^T \in {X_T}\label{eq11}
	\end{equation}
	
	We take $\eta $ as a threshold for dividing easy and hard samples. A list of easy and hard samples is obtained according to Eq.(\ref{eq12}) and Eq.(\ref{eq13}). \par
	\begin{equation}
		hard\_list = \{ x_i^T|H(x_i^T) > \eta \} \label{eq12}
	\end{equation}
	\begin{equation}
		easy\_list = \{ x_i^T|H(x_i^T) \le \eta \} \label{eq13}
	\end{equation}
	
	\noindent{\textbf{Global Pseudo-label Correction Based on Class Prototype:}}
	In this step, the pseudo-label is corrected by considering the relative distance between the feature of each pixel and the prototypes of the foreground and background classes.\par
	Firstly, we gain the segmentation prediction $\hat y_i^{S \rightarrow T}(j)$ and the feature vector $f_i^T(j)$ of each pixel by taking the target image $x_i^T$ as input into the model. Secondly, we calculate the foreground and background class prototype vectors of easy samples according to Eq.(\ref{eq4}) and Eq.(\ref{eq5}),\par
	\begin{equation}
		{f_{bck}} = \sum\limits_{i = 0}^{{N_S}} {\sum\nolimits_j {\mathbbm{1}[p_i^{S \rightarrow T}(j) =  = 0]*f_i^T(j)} } /\sum\limits_{i = 0}^{{N_S}} {\sum\nolimits_j {\mathbbm{1}[} p_i^{S \rightarrow T}(j) =  = 0]} \label{eq4}
	\end{equation}
	\begin{equation}
		{f_{obj}} = \sum\limits_{i = 0}^{{N_S}} {\sum\nolimits_j {\mathbbm{1}[p_i^{S \rightarrow T}(j) =  = 1]*f_i^T(j)} } /\sum\limits_{i = 0}^{{N_S}} {\sum\nolimits_j {\mathbbm{1}[} p_i^{S \rightarrow T}(j) =  = 1]} \label{eq5}
	\end{equation}
	where ${f_{bck}}$ represents the background  prototype vector of easy samples, and ${f_{obj}}$ is the foreground prototype vector of easy samples. $N_S$ is the number of easy samples in a batch. $f_{i}^T(j)$ is the feature vector of pixel $j$.
	Thirdly, we calculate the distance between the feature vector of each pixel $j$ and the prototype vectors ${f_{bck}}$ and ${f_{obj}}$ by inner product as Eq.(\ref{eq6}) and Eq.(\ref{eq7}).  Fourthly, we get the mask based on its relative distance as Eq.(\ref{eq8}),\par
	\begin{equation}
		{d_{obj,i}(j)} = ||{f_{obj}},f_i^T(j)|{|_2}\label{eq6}
	\end{equation}
	\begin{equation}
		{d_{bck,i}(j)} = ||{f_{bck}},f_i^T(j)|{|_2}\label{eq7}
	\end{equation}
	\begin{equation}
		mas{k_{ic}} = \left\{ {\begin{array}{*{20}{c}}
				{1,{d_{obj,i}} \ge {d_{bck,i}}}\\
				{0,{d_{obj,i}} < {d_{bck,i}}}
		\end{array}} \right.\label{eq8}
	\end{equation}
	where $|| \cdot ||$ means 2-norm. The value of c is 0 or 1, and we get a binary optic cup mask $mask_{i0}$ and a binary optic disc $mask_{i1}$, respectively in our experiment.\par
	
	Lastly, we initialize a zeros matrix ${p_i^{T}} $ with the same size as $p_i^{S \rightarrow T}$, namely ${p_i^{T}} \in {^{{H_2} \times {W_2} \times {C_2}}}$, where ${H_2}$ and ${W_2}$ represent the height and width of the pseudo-label respectively; ${C_2}$ represents the number of segmentation categories. Set the pseudo-label to 1 only when the corrected pseudo-label is consistent with the mask prediction, that is, the position is a trusted foreground prediction,\par
	\begin{equation}
		{p_i^{T}} = \left\{ {\begin{array}{*{20}{c}}
				{1,p_i^{S \rightarrow T} = mask_i}\\
				{0,else}
		\end{array}}\label{eq17} \right.
	\end{equation}
	In Eq.(\ref{eq17}) ${p_i}$ denotes the local-global corrected pseudo-label. In the process of network adaptive training, with the update of network parameters, $f_{i}^T$ is constantly updated, and the prototypes of easy samples are also updated online with network training.
	
	As the model gradually adapts to the data distribution of the target domain, the target and background class prototypes of easy samples are gradually updated to more realistic class prototype vectors in the target domain. In the network iteration process, the updated class prototype vector is used for global pseudo-label correction, and the online updated corrected pseudo-label ${p_i}$ is used as the label of the target domain sample for source-free domain adaptative training.\par
	
	\subsection{loss function} 
	We use pseudo-labels to optimize the target domain training model. The overall loss function of the proposed method (LGDA) is formulated as Eq.(\ref{eq16}),\par
	\begin{equation}
		{L_T} =  - \sum\limits_{x_i^T \in {X_T}} {{p_i^{T}}\log ({\phi _T}(x_i^T))}  + (1 - {p_i^{T}})(1 - \log ({\phi _T}(x_i^T)) \label{eq16}
	\end{equation}
	where ${\phi _T}(x_i^T)$ is the prediction of the target domain model, and ${\phi _T}(x_i^T)$ is the pseudo-label after local-global pseudo-label optimization.
	
	\section{Experiment Settings}
	\label{section4}
	\subsection{Dataset and evaluation metric}
	We use three datasets from different sites to validate the effectiveness of the proposed LGDA. REFUGE challenge\cite{orlando2020refuge} dataset is regarded as the source domain and RIM-ONE-r3\cite{fumero2011rim} and Drishti-GS\cite{sivaswamy2015comprehensive} datasets are treaded as target domains. The source domain includes 400 annotated training images, and the two target domain data are split into 99/60 and 50/51 images for training/testing respectively. The data preprocessing of this paper follows the setting in the literature\cite{chen2021source}. The fundus image is cropped into a region of interest (ROI) centered on the optic disc as the network input with the size of 512$\times$512. Additionally, we use the common data augmentation strategies including random rotation, flipping, elastic transformation, contrast adjustment, adding Gaussian noise, and random erasing. For evaluation, we employ two commonly used metrics, including the Dice coefficient for overlap measurement and average surface distance (ASD) for boundary consistency evaluation. Higher Dice and lower ASD indicate better performance. \par
	
	\subsection{Implementation details}
	In this study, we employ the DeepLabv3+ network as the backbone segmentation network. All methods are implemented in PyTorch and trained on one NVIDIA TITAN RTX GPU. The batch size is set to 8. We use the Adam optimizer in our experiments. We set the fixed learning rate to 0.001. We set N to 3 and K to 5 in local denoising moudel. All experiments follow the same training settings for a fair comparison.

	\section{Experimental results and analysis}
	\label{section5}
	\subsection{Comparison with State-of-the-Arts}
	We compare our method with recent state-of-the-art domain adaptation methods, including BEAL\cite{wang2019boundary},AdvEnt\cite{vu2019advent},DAE\cite{karani2021test},SRDA\cite{bateson2020source}, TT\cite{vs2022target}, FSM\cite{yang2022source} and  DPL\cite{chen2021source}. Among them, the BEAL\cite{wang2019boundary} and AdvEnt\cite{vu2019advent} with the access of source domain. DAE\cite{karani2021test} , SRDA\cite{bateson2020source}, TT \cite{vs2022target}, FSM\cite{yang2022source} and DPL\cite{chen2021source} are source-free domain adaptation methods, but DAE\cite{karani2021test} and SRDA\cite{bateson2020source} introduce auxiliary networks to predict prior information during source domain pre-training. In Table \ref{RIM-ONE-r3数据集LGDA} and Table \ref{Drishti-GS数据集LGDA}, "w/o adaptation" represents the results obtained by directly testing the source domain model on the target domain test set, while "Upper bound" denotes the results achieved by training a model directly on the target domain training set and subsequently testing it. Additionally, we denote the unsupervised domain adaptation method as "U", the source-free domain adaptation method as "F", and the source-free test-time DA as "T". \par
	
	Table \ref{RIM-ONE-r3数据集LGDA} and Table \ref{Drishti-GS数据集LGDA} report the quantitative experimental results on the RIM-ONE-r3 dataset and Drishti GS dataset, respectively. 
	\par
	\begin{table}[H]
		\renewcommand{\arraystretch}{1.2}
		\centering
		\caption{Comparison of experimental results of LGDA model in RIM-ONE-r3 dataset}
		\label{RIM-ONE-r3数据集LGDA}
		\resizebox{\linewidth}{!}{{\begin{tabular}{ccccccc}
					\hline
					\multirow{2}{*}{Method} & \multicolumn{2}{c}{Optic disc segmentation} & \multicolumn{2}{c}{Optic cup segmentation} & \multicolumn{2}{c}{Avg}\\
					& Dice $\uparrow$ $[\% ]$ & ASD $\downarrow$ (pixel) & Dice $\uparrow$ $[\% ]$ & ASD $\downarrow$(pixel)& Dice $\uparrow$ $[\% ]$ & ASD $\downarrow$ (pixel)\\
					\hline
					w \textbackslash o adaptation  & 88.69$ \pm $0.04  & 10.63$ \pm $4.14 & 75.06$ \pm $25.09 & 8.36$ \pm $5.55 & 81.87 & 9.50 \\
					Upper bound   & 94.74$ \pm $2.16  & 4.43$ \pm $1.70  & 80.58$ \pm $20.92 & 7.06$ \pm $6.83  & 87.66 & 5.75  \\
					\hline
					BEAL\cite{wang2019boundary}(U)     & 88.70$ \pm $3.53  & 16.63$ \pm $5.58  & 79.00$ \pm $12.29 & 14.49$ \pm $6.78  & 83.85 & 15.56 \\
					AdvEnt\cite{vu2019advent}(U)   & 89.73$ \pm $3.66  & 9.84$ \pm $3.86  & 77.99$ \pm $21.08 & 7.57$ \pm $4.24   & 83.86 & 8.71  \\
					\hline
					DAE\cite{karani2021test}(T)      & 89.09$ \pm $3.32  & 11.63$ \pm $6.84  & 79.01$ \pm $12.82 & 10.31$ \pm $8.45  & 84.05 & 10.97 \\
					SRDA\cite{bateson2020source}(F)     & 89.37$ \pm $2.70  & 9.91$ \pm $2.45   & 77.61$ \pm $13.58 & 10.15$ \pm $5.75  & 83.49 & 10.03 \\
					TT\cite{vs2022target}(F) & 65.34$ \pm $25.91 & 26.30$ \pm $24.75 & 44.87$ \pm $29.31 & 22.28$ \pm $17.74 & 55.10 & 24.29 \\
					FSM\cite{yang2022source}(F) & 84.42$ \pm $4.19 & 16.53$ \pm $9.44 & 80.14$ \pm $13.28 & 8.33$ \pm $4.70 & 82.28 & 12.43 \\
					DPL\cite{chen2021source}(F)      & 89.47$ \pm $4.56  & \textbf{6.92$ \pm $8.24}   & 81.93$ \pm $14.96 & 9.56$ \pm $3.57   & 85.70 & 8.24  \\
					LGDA(Ours)(F)     & \textbf{91.15$ \pm $2.63}  & 7.00$ \pm $3.34   & \textbf{82.64$ \pm $10.57} & \textbf{8.33$ \pm $2.87}   & \textbf{86.89} & \textbf{7.66} \\ \hline
			\end{tabular}}
		}
		
	\end{table}
	
	According to Table \ref{RIM-ONE-r3数据集LGDA}, performance on both average Dice coefficient and average ASD in RIM-ONE-r3 dataset are 86.69\% and 7.66, respectively. The highest Dice score and lowest ASD on the RIM-ONE-r3 dataset indicate the proposed method can effectively make the segmentation model pre-trained in the source domain adaptive in the RIM-ONE-r3 dataset under the source-free condition. For optic disc segmentation, LGDA achieves 91.15\% on Dice, which is 1.68\% higher than the second-best DPL's performance. Additionally, for optic cup segmentation, LGDA achieves 82.64\% on Dice, which is 0.71\% higher than the second-best DPL's performance. These results demonstrate that the local-global pseudo-label correction method effectively corrects the incorrectly predicted pseudo-labels in the target domain so that the model can better adapt to the data distribution of the target domain. Our method achieves the second-best ASD of 7.00 on the optic disc segmentation task, which has a very close performance with the best ASD of 6.92 achieved by DPL\cite{chen2021source}. The ASD achieved by our model for cup segmentation is lower than other comparative methods, indicating that our method can better preserve the boundaries for optic cup segmentation. It is important to highlight that our model's segmentation results outperform BEAL and Advent in terms of Dice and ASD metrics. This indicates that our method achieves superior performance even without access to the source dataset, surpassing some popular UDA methods. The success of our approach attribut to the direct adaptation of the source domain pre-trained model to the target domain data through our proposed LGDA model. This adaptation enables the model to capture more discriminative feature representations, which in turn corrects the noisy components present in the original pseudo labels for source-free domain adaptation segmentation. In contrast, unsupervised domain adaptation methods face challenges in effectively eliminating domain differences and acquiring domain invariant features.\par
	
	\begin{figure}[h]
		\centering\includegraphics[scale=0.6]{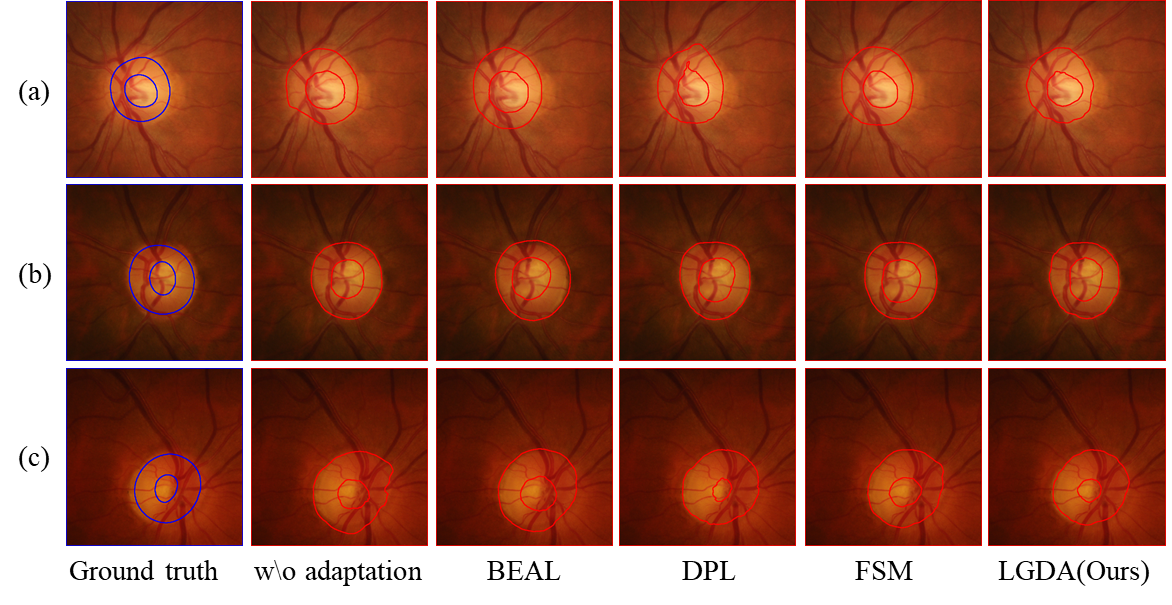}\caption{Visual segmentation results for typical samples on the RIM-ONE-r3 dataset}\label{visR}
	\end{figure}
	
	Figure \ref{visR} depicts the qualitative segmentation results of some typical samples on the RIM-ONE-r3
	dataset. We can observe that our method generates more consistent optic disc segmentation
	and optic cup segmentation results with ground truth than other methods. The accurate and stable predictions validate the reliability of our method to face source-free adaptation.\par
		\begin{table}[H]
		\renewcommand{\arraystretch}{1.2}
		\centering
		\caption{Comparison of experimental results of LGDA model in Drishti GS dataset}
		\label{Drishti-GS数据集LGDA}
		\resizebox{\linewidth}{!}{{	\begin{tabular}{ccccccc}
					\hline
					\multirow{2}{*}{Method} & \multicolumn{2}{c}{Optic disc segmentation} & \multicolumn{2}{c}{Optic cup segmentation} & \multicolumn{2}{c}{Avg}\\
					& Dice $\uparrow$ $[\% ]$ & ASD $\downarrow$ (pixel) & Dice $\uparrow$ $[\% ]$ & ASD $\downarrow$(pixel)& Dice $\uparrow$ $[\% ]$ & ASD $\downarrow$ (pixel)\\
					\hline
					w \textbackslash adaptation  & 96.66$ \pm $1.12 & 3.78$ \pm $1.34 & 81.55$ \pm $11.94 & 11.94$ \pm $7.86 & 89.10 & 7.86 \\
					Upper bound   & 96.65$ \pm $1.60  & 3.60$ \pm $1.50  & 89.09$ \pm $11.23 & 6.78$ \pm $3.68  & 92.87 & 5.19  \\
					\hline
					BEAL\cite{wang2019boundary}(U)    & 95.54$ \pm $2.09  & 7.78$ \pm $3.37  & 85.95$ \pm $11.44 & 14.51$ \pm $8.15  & 90.75 & 11.14 \\
					AdvEnt\cite{vu2019advent}(U)   & 96.16$ \pm $1.65  & 4.36$ \pm $1.83  & 82.75$ \pm $11.08 & 11.36$ \pm $7.22  & 89.46 & 7.86  \\
					\hline
					DAE\cite{karani2021test}(T)      & 94.04$ \pm $2.85  & 8.79$ \pm $7.45  & 83.11$ \pm $11.89 & 11.56$ \pm $6.32  & 88.58 & 10.18 \\
					SRDA\cite{bateson2020source}(F)     & 96.22$ \pm $1.30  & 4.88$ \pm $3.47  & 80.67$ \pm $11.78 & 13.12$ \pm $6.48  & 88.45 & 9.00  \\
					TT\cite{vs2022target}(F) & 86.59$ \pm $12.79 & 14.32$ \pm $8.53 & 61.64$ \pm $25.51 & 20.14$ \pm $10.63 & 74.15 & 17.23 \\
					FSM\cite{yang2022source}(F) & 95.85$ \pm $2.36 & 4.67$ \pm $2.47 & 82.24$ \pm $13.30 & 12.03$ \pm $6.56 & 89.04 & 8.35 \\
					DPL\cite{chen2021source}(F)      & 96.53$ \pm $1.29  & 3.92$ \pm $1.43  & 83.15$ \pm $11.78 & 11.42$ \pm $6.56  & 89.84 & 7.67  \\
					LGDA(Ours)(F)     & \textbf{96.86$ \pm $1.23}  & \textbf{3.50$ \pm $1.28}  & \textbf{86.10$ \pm $10.67} & \textbf{9.13$ \pm $5.18}   & \textbf{91.48} & \textbf{6.32}  \\ \hline
			\end{tabular}}
		}
	\end{table}
	In the Drishti-GS dataset, our LGDA method achieves the best Dice and ASD scores for optic segmentation and optic cup segmentation, as reported in Table \ref{Drishti-GS数据集LGDA}. Specifically, our method achieves an average Dice score of 91.48\%, which is higher than the second-best method by a significant margin, and it performs 6.32\% lower than the second-best method on ASD. For optic disc segmentation, LGDA achieves a Dice score of 96.86\%, surpassing the second-best method DPL by 0.33\%. Additionally, in optic cup segmentation, our method achieves a Dice score of 86.10\%, which is 2.95\% higher than the performance of DPL. These results clearly demonstrate the effectiveness of our local-global pseudo-label correction approach in rectifying inaccurately predicted pseudo-labels within the target domain. This correction process enhances the model's ability to align with the data distribution of the target domain. Simultaneously, the segmentation results of our model outperform BEAL and Advent on Dice and ASD, suggesting that our method outperforms some popular UDA methods without access to the source dataset.\par

\begin{figure}[h]
		\centering\includegraphics[scale=0.6]{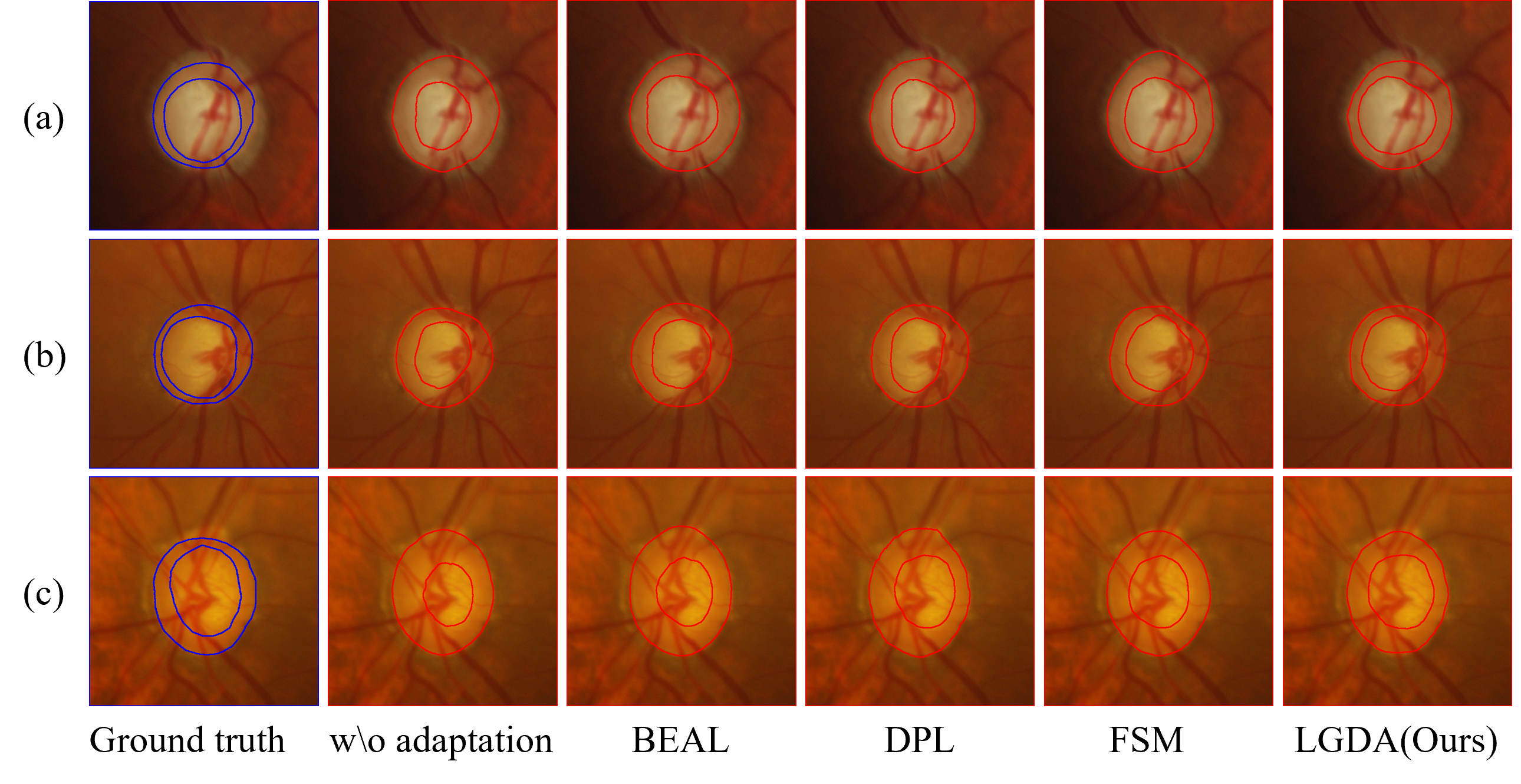}\caption{Visual segmentation results for typical samples on the Drishti-GS dataset}\label{visD}
	\end{figure}

	Figure \ref{visD} shows some qualitative evaluation results on Drishti-GS. Although FSM[12] and DPL[9] outperformed the other existing SFDA methods, they have some visible errors, because they fail to fully leverage the local information present in the target domain images. The results of our method have the highest consistency with the ground truth, indicating the effectiveness and robustness of our method.
	
	\subsection{Ablation Study}
	\subsubsection{ Ablation Studies on Easy and hard sample division}
	
	\paragraph{\textbf{The method of easy and hard sample division}}
	
	We conduct experiments utilizing three distinct methods to divide the target domain data into simple and hard samples, and mark them as "TopK," "AVG," and "H \textgreater{} $\eta $," respectively. In the first method, we select the TopK samples in the target domain dataset where K is set to 0.3 times the total number of samples in the training set. In the second method, we calculate the average entropy value across the entire target domain training and then choose samples with entropy values lower than the calculated average as simple samples. As for the third method, we designate samples with values lower than the threshold $\eta $ as simple samples. In this case, we assume that the prediction probability of each pixel exceeds 0.9, which is considered more reliable. The entropy value of the entire image is then calculated using Eq.\ref{entropy}. In our implementation, we set $\eta $ to 0.044.\par
	\begin{table}[H]
		\renewcommand{\arraystretch}{1}
		\centering
		\caption{The validation of easy and hard sample partitioning of the RIM-ONE-r3 dataset}
		\label{RIM-ONE-r3阈值}
		\begin{tabular}{cccc}
			\hline
			& Optic disc$[\% ]$  & Optic cup$[\% ]$  & Avg$[\% ]$   \\ 
			\hline
			TopK                    & 91.10       & 81.39      & 86.24 \\
			AVG                             & 93.76        & 78.84     & 86.30 \\
			H\textgreater{}$\eta $     & \textbf{91.15} & \textbf{82.64} & \textbf{86.89} \\ \hline
		\end{tabular}
	\end{table}
	\begin{table}[H]
		\renewcommand{\arraystretch}{1}
		\centering
		\caption{The validation of easy and hard sample partitioning of the Drishti-GS dataset}
		\label{Drishti-GS阈值}
		
		\begin{tabular}{cccc}
			\hline
			& Optic disc$[\% ]$  & Optic cup$[\% ]$  & Avg$[\% ]$   \\ 
			\hline
			TopK                   & 96.31 & 82.51 & 89.41 \\
			AVG                            & 94.61 & 79.46 & 87.04 \\
			H\textgreater{}$\eta $     & \textbf{96.86} & \textbf{86.10} & \textbf{91.48} \\ \hline
		\end{tabular}
	\end{table}
	
	The experimental results are presented in Table \ref{RIM-ONE-r3阈值} and Table \ref{Drishti-GS阈值}. The Dice coefficient is employed as the evaluation metric. By comparing the experimental outcomes on both datasets, we observe that the third method is more suitable for dividing samples into easy and hard categories. Consequently, we select the third method for dividing samples into simple and difficult categories in the final model. \par 
	
	\begin{figure}[h]
		\centering\includegraphics[scale=0.27]{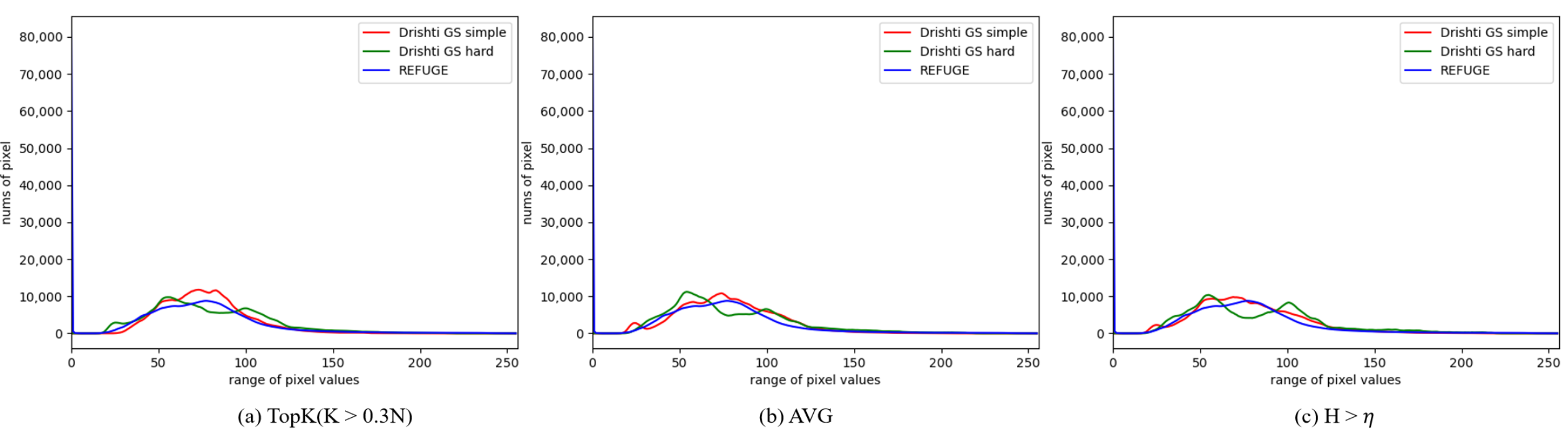}\caption{Drishti-GS dataset easy difficult sample segmentation grayscale statistics}\label{shD1}
	\end{figure}
	

	To validate the effectiveness of the selected method for dividing samples into easy and hard categories, we conducte experiments using the three aforementioned methods. Firstly, we apply these methods to obtain two lists: one comprising easy samples and the other consisting of hard samples. Next, we analyze the grayscale distribution of the grayscale images associated with the easy and hard samples, respectively. Finally, we compare these distributions with the grayscale distribution of the source domain data image. Figure \ref{shD1} illustrates the statistical results of the grayscale distribution, where each graph represents the outcomes obtained through the "TopK," "AVG," and "H\textgreater{}$\eta $" division methods, respectively. The blue, red, and green lines correspond to the gray distribution of the source domain dataset, easy samples of the target domain dataset, and difficult samples of the target domain dataset, respectively. By examining the statistical results, it is evident that the gray distribution of the simple samples divided using the "H\textgreater{}$\eta $" method closely aligns with the gray distribution of the source domain. This observation confirms the rationale behind selecting the third-division method as the more reasonable approach.
	
	\paragraph{\textbf{Impact of dividing easy and hard sample on global denoising module}}
	
	We conduct two sets of experiments to verify the impact dividing easy and hard sample on global denoising module. These two sets of experiments are the LGDA method without using easy sample class prototypes for global denoising and the LGDA method using easy samples class prototypes for global denoising.The experimental results are shown in Table \ref{isDR} and Table \ref{isDD}. 
	\begin{table}[H]
		\renewcommand{\arraystretch}{1.2}
		\centering
		\caption{Impact of dividing easy and hard sample on global denoising module in RIM-ONE-r3 dataset}
		\label{isDR}
		\resizebox{\linewidth}{!}{{\begin{tabular}{ccccccc}
					\hline
					\multirow{2}{*}{Method} & \multicolumn{2}{c}{Optic disc segmentation} & \multicolumn{2}{c}{Optic cup segmentation} & \multicolumn{2}{c}{Avg}\\
					& Dice $[\% ]$ & ASD(piexl) & Dice $[\% ]$ & ASD(piexl)& Dice $[\% ]$ & ASD(piexl)\\
					\hline
					w \textbackslash o division   & 89.93$ \pm $9.27  & 9.27$ \pm $3.04  & 76.60$ \pm $7.90  & 7.90$ \pm $5.05 & 83.27 & 8.58 \\
					w division & \textbf{91.15$ \pm $2.63}  & \textbf{8.33$ \pm $2.87}  & \textbf{82.64$ \pm $10.57} & \textbf{7.00$ \pm $3.34} & \textbf{86.89} & \textbf{3.11} \\ \hline
				\end{tabular}
			}
		}
	\end{table}
	\begin{table}[H]
	\renewcommand{\arraystretch}{1.2}
	\centering
	\caption{Impact of dividing easy and hard sample on global denoising module in Drishti-GS dataset}
	\label{isDD}
	\resizebox{\linewidth}{!}{{\begin{tabular}{ccccccc}
				\hline
				\multirow{2}{*}{Method} & \multicolumn{2}{c}{Optic disc segmentation} & \multicolumn{2}{c}{Optic disc segmentation} & \multicolumn{2}{c}{Avg}\\
				& Dice $[\% ]$ & ASD(piexl) & Dice $[\% ]$ & ASD(piexl)& Dice $[\% ]$ & ASD(piexl)\\
				\hline
				w \textbackslash o division & 96.82$ \pm $1.34 & 3.57$ \pm $1.45 & 83.22$ \pm $10.16 & 11.07$ \pm $6.29 & 90.02 & 7.32 \\
				w division     & \textbf{96.86$ \pm $1.23} & \textbf{3.50$ \pm $1.28} & \textbf{86.10$ \pm $10.67} & \textbf{9.13$ \pm $5.18}  & \textbf{91.48} & \textbf{6.32} \\ \hline
		\end{tabular}}
	}
	\end{table}

	Compared to using a simple easy class prototype for denoising methods, using all sample class prototypes for denoising methods results in a significant decrease in the Dice coefficient for optic disc and optic cup segmentation. This attributs to the presence of biased prototypes. These biased class prototypes cause the optic cup pixels to be closer to the prototype of the background class. By incorporating a global pseudo-label correction method based on easy sample, the performance of optic disc segmentation is enhanced. Specifically, the Dice coefficients for the RIM-ONE-r3 and Drishti-GS datasets show improvements of 1.22\% and 0.04\% respectively. In terms of optic cup segmentation, our method demonstrates a significant enhancement in Dice by 6.04\% and 2.88\% compared to the method that does not divide samples into simple ones and hard ones. These results highlight the importance of extracting unbiased prototype vectors in order to improve the accuracy of segmentation.
	
	\subsubsection{Ablation Studies on Key Components}
	We conduct ablation experiments on the REFUGE dataset and Drishti-GS dataset to validate the effectiveness of each module of our proposed model. Concretely, we use the test results generated directly from the source domain pre-trained model as the Baseline. Next, we incorporated a local pseudo-label correction module onto the baseline method, denoted as ``Baseline+L". Finally, we intorduce global denoising module based on easy sample class prototype, denoted as ``Baseline+LG"(i.e., the final model LGDA). We use the Dice coefficient and the ASD coefficient as evaluation metrics for ablation experiments. \par

	\begin{table}[H]
		\renewcommand{\arraystretch}{1.2}
		\centering
		\caption{Ablation experiment results of LGDA model in RIM-ONE-r3 dataset}
		\label{RIM-ONE-r3消融实验}
		\resizebox{\linewidth}{!}{{\begin{tabular}{ccccccc}
					\hline
					\multirow{2}{*}{Method} & \multicolumn{2}{c}{Optic disc segmentation} & \multicolumn{2}{c}{Optic cup segmentation} & \multicolumn{2}{c}{Avg}\\
					& Dice $[\% ]$ & ASD(piexl) & Dice $[\% ]$ & ASD(piexl)& Dice $[\% ]$ & ASD(piexl)\\
					\hline
					Baseline         & 88.69$ \pm $0.04  & 10.63$ \pm $4.14 & 75.06$ \pm $25.09 & 8.36$ \pm $5.55 & 81.87 & 9.50 \\
					Baseline+L      & 88.91$ \pm $10.48 & 10.48$ \pm $3.68 & 80.91$ \pm $7.71  & 7.71$ \pm $6.65 & 84.94 & 9.09 \\
					Baseline+LG & \textbf{91.15$ \pm $2.63}  & \textbf{8.33$ \pm $2.87}  & \textbf{82.64$ \pm $10.57} & \textbf{7.00$ \pm $3.34} & \textbf{86.89} & \textbf{3.11} \\ \hline
				\end{tabular}
			}
		}
	\end{table}
	
	\begin{table}[H]
		\renewcommand{\arraystretch}{1.2}
		\centering
		\caption{Ablation experiment results of LGDA model in Drishti-GS dataset}
		\label{Drishti-GS消融实验}
		\resizebox{\linewidth}{!}{{\begin{tabular}{ccccccc}
					\hline
					\multirow{2}{*}{Method} & \multicolumn{2}{c}{Optic disc segmentation} & \multicolumn{2}{c}{Optic disc segmentation} & \multicolumn{2}{c}{Avg}\\
					& Dice $[\% ]$ & ASD(piexl) & Dice $[\% ]$ & ASD(piexl)& Dice $[\% ]$ & ASD(piexl)\\
					\hline
					Baseline       & 96.66$ \pm $1.12 & 3.78$ \pm $1.34 & 81.55$ \pm $11.94 & 11.94$ \pm $7.86 & 89.10 & 7.86 \\
					Baseline+L    & 96.75$ \pm $1.40 & 3.68$ \pm $1.50 & 83.37$ \pm $10.28 & 10.98$ \pm $6.26 & 90.06 & 7.33 \\
					Baseline+LG    & \textbf{96.86$ \pm $1.23} & \textbf{3.50$ \pm $1.28} & \textbf{86.10$ \pm $10.67} & \textbf{9.13$ \pm $5.18}  & \textbf{91.48} & \textbf{6.32} \\ \hline
			\end{tabular}}
		}
	\end{table}
	
	The experimental results are shown in Table \ref{RIM-ONE-r3消融实验} and Table \ref{Drishti-GS消融实验}. Firstly, after incorporating the local denoising module, the optic disc segmentation achieves Dice coefficients of 88.91\% and 97.75\% on the RIM-ONE-r3 dataset and Drishti-GS dataset, respectively. Compared to the baseline, there is an improvement of 0.22\% and 0.09\%, respectively. These results indicate that the local pseudo-label correction module remains effective even in the presence of small segmentation errors. Additionally, for the optic cup segmentation task, the Dice coefficients reach 80.91\% and 83.37\%, representing an improvement of 5.85\% and 1.82\% compared to the baseline. The significant enhancement in optic cup segmentation demonstrates that the local denoising module effectively leverages local spatial continuity to refine pixel-level segmentation, thereby providing reliable initial pseudo-labels for adaptive training.
	
	Furthermore, the integration of the class prototype-based global pseudo-label correction brings a significant improvment in optic disc segmentation accuracy across both target domains. Specifically, after introducing the global denoising module, the Dice scores for the optic cup segmentation task on the RIM-ONE-r3 dataset increased by 2.24\% and 2.03\% respectively. At the same time, the ASD coefficients decreased by 2.15 and 0.71. On the Drishti-GS dataset, the Dice scores for the optic disc segmentation and optic cup segmentation tasks increased by 0.11\% and 2.73\% respectively, while the ASD scores decreased by 0.18 and 1.85. The global denoising moudle utilizing class prototypes leverages the global image information to rectify pseudo-labels assigned to pixels that exhibit higher similarity to the target class prototype in the feature space but are mistakenly labeled as background. 
	
	
	Ultimately, the optimal performance of our model is achieved through the integration of the local denoising and global denoising modules.  The superior performance demonstrates the effectiveness of our approach, which utilizes source domain prior knowledge to effectively leverage the local spatial continuity and global semantic information of target domain images during training. By employing offline correction of the local pseudo-labels , it provides reliable initial pseudo-labels for adaptive training and effectively prevents error accumulation. And by employing online correction of pseudo-labels using the global denoising module, the accuracy of pseudo-labels is further enhanced, enabling the model to successfully adapt to the data distribution of the target domain.
	
	\begin{figure}[h]
		\centering\includegraphics[scale=0.6]{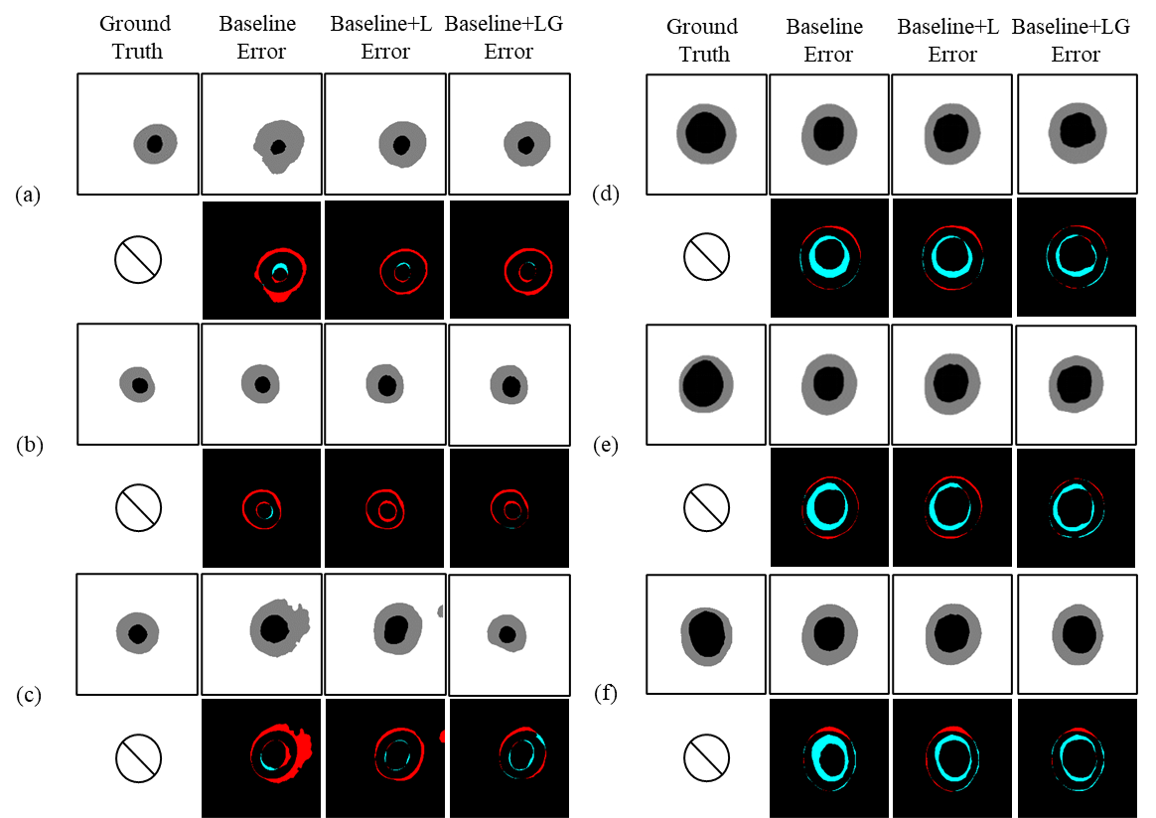}\caption{LGDA model ablation experiment error visualization results on RIM-ONE-r3 dataset}\label{Rerror}
	\end{figure} 
	
	The segmentation error visualization results of some typical samples on the RIM-ONE-r3 dataset and Drishti-GS dataset are shown in Figure \ref{Rerror}. In Figure \ref{Rerror}, each sample is represented by two rows. The first row visually presents the segmentation results, while the second row depicts the errors in comparison to the ground truth. The first column represents the ground truth, and the subsequent columns (second to fourth) correspond to the results of the Baseline, Baseline + L, and Baseline + LG, respectively. Within the error maps, the blue regions indicate areas where the predictions fall short of the ground truth, while the red regions highlight areas where the predictions exceed the ground truth.
	
	By comparing the second and fourth columns, we can observe a significant reduction in errors for the LGDA model relative to the baseline model. Specifically, in examples (a) and (c), a distinct decrease in prediction errors is evident when incorporating local pseudo-label correction in the adaptive model, as seen from the comparison between the second and third columns. After partitioning the target domain samples into easy ones and hard ones and subsequently applying global pseudo-label correction, in example (c), there is a notable enhancement in segmentation quality, surpassing the results obtained from local pseudo-label correction. This observation underscores the effectiveness of the pseudo-label correction method based on prototypes derived from easy samples following the introduction of easy and hard sample classification.
	
	Moreover, through a step-by-step visualization and comparison of the errors introduced by each module, it is apparent that the model's prediction errors steadily decrease. This observation provides compelling evidence for the interplay and complementary nature of the various modules proposed, ultimately resulting in a notable improvement in the model's performance for source-free domain adaptation in the optic cup and disc segmentation tasks.

	\section{Conclusions}
	\label{section7}
	In this paper, we propose a local-global pseudo-label correction method for reliable source-free domain adaptation in the medical image segmentation method. Our LGDA can provide more accurate pseudo-labels for the target domain to improve the segmentation performance of the source domain model in the target domain dataset. We propose a local label correction method that can take the neighborhood pixels into consideration. Second, we propose a global label correction method from the perspective of prototypical learning. To reduce errors in a global pseudo-label correction, we propose an easy-hard sample division method. Extensive experiments on RIM-ONE-r3 and Drishti-GS demonstrate the effectiveness of our method LGDA. In the RIM-ONE-r3 dataset, LGDA can improve 2.46\% and 7.58\% on Dice in the optic disc and cup segmentation compared to the results generated directly from the source domain pre-trained model. In the Drishti-GS dataset, LGDA can improve 0.20\% and 4.55\% on Dice in the optic disc and cup segmentation compared to baseline. Our forthcoming research will concentrate on the challenges of federated learning in the field of medical segmentation. Furthermore, we will explore the issue of federative learning in the context of weak supervision.
	

	\bibliography{mybibfile}

\end{document}